\pdfoutput=1

\documentclass[11pt]{article}

\usepackage[final]{acl}

\usepackage{times}
\usepackage{latexsym}

\usepackage[T1]{fontenc}

\usepackage[utf8]{inputenc}

\usepackage{microtype}

\usepackage{inconsolata}
\usepackage{enumitem}
\usepackage{graphicx}
\usepackage{tikz}
\usepackage{adjustbox}
\usepackage{booktabs}
\usepackage{amssymb}
\usepackage{siunitx}
\usepackage{tabularx}
\usepackage{colortbl}
\usepackage{comment}

\usepackage{multirow}
\usepackage[table]{xcolor}

\newcolumntype{R}[1]{>{\raggedleft\arraybackslash}p{#1}}

\usepackage{xspace}
\newcommand{\datasetName}{BabyBabelLM\xspace}
\newcommand{\numLanguages}{45\xspace}

\newcommand{\numTierOne}
{9\xspace}

\newcommand{\numTierTwo}{15\xspace}

\newcommand{\numTierThree}{21\xspace}

\newcommand{\cready}[1]{}
\newcommand{\TODO}[1]{\textcolor{blue}{\small [TODO: #1]}}

\title{BabyBabelLM:\\A Multilingual Benchmark of Developmentally Plausible Training Data} 
\author{
\parbox{\textwidth}{\centering
 \textbf{Jaap Jumelet}\textsuperscript{1}, \textbf{Abdellah Fourtassi}\textsuperscript{2}, \textbf{Akari Haga}\textsuperscript{3}, \textbf{Bastian Bunzeck}\textsuperscript{4}, \textbf{Bhargav Shandilya}\textsuperscript{5}, \textbf{Diana Galvan-Sosa}\textsuperscript{6, 12}, \textbf{Faiz Ghifari Haznitrama}\textsuperscript{7}, \textbf{Francesca Padovani}\textsuperscript{1}, \textbf{Francois Meyer}\textsuperscript{8}, \textbf{Hai Hu}\textsuperscript{9}, \textbf{Julen Etxaniz}\textsuperscript{10}, \textbf{Laurent Prévot}\textsuperscript{2}, \textbf{Linyang He}\textsuperscript{11}, \textbf{María Grandury}\textsuperscript{12}, \textbf{Mila Marcheva}\textsuperscript{6}, \textbf{Negar Foroutan}\textsuperscript{13}, \textbf{Nikitas Theodoropoulos}\textsuperscript{14}, \textbf{Pouya Sadeghi}\textsuperscript{15}, \textbf{Siyuan Song}\textsuperscript{16}, \textbf{Suchir Salhan}\textsuperscript{6}, \textbf{Susana Zhou}\textsuperscript{12}, \textbf{Yurii Paniv}\textsuperscript{17}, \textbf{Ziyin Zhang}\textsuperscript{18}, \textbf{Arianna Bisazza}\textsuperscript{1}, \textbf{Alex Warstadt}\textsuperscript{19}, \textbf{Leshem Choshen}\textsuperscript{20}\\
 }
\\
\\
\parbox{\textwidth}{\centering\footnotesize
 \textsuperscript{1}University of Groningen,
\textsuperscript{2}Aix Marseille University,
\textsuperscript{3}Nara Institute of Science and Technology,
\textsuperscript{4}Bielefeld University,
\textsuperscript{5}University of Colorado Boulder,
\textsuperscript{6}University of Cambridge,
\textsuperscript{7}KAIST,
\textsuperscript{8}University of Cape Town,
\textsuperscript{9}City University of Hong Kong,
\textsuperscript{10}HiTZ, University of the Basque Country,
\textsuperscript{11}Columbia University,
\textsuperscript{12}SomosNLP,
\textsuperscript{13}EPFL,
\textsuperscript{14}Independent Researcher,
\textsuperscript{15}University of Tehran,
\textsuperscript{16}University of Texas at Austin,
\textsuperscript{17}Ukrainian Catholic University,
\textsuperscript{18}Shanghai Jiao Tong University,
\textsuperscript{19}University of California San Diego,
\textsuperscript{20}MIT, MIT-IBM Watson AI Lab \\\vspace*{5pt}Correspondence to 
 \url{j.w.d.jumelet@rug.nl} and \url{leshem.choshen@mail.huji.ac.il}\thanks{Author contributions provided in Appendix~\ref{sec:contributions}.}
}}

\begin{document}
 \maketitle
\begin{abstract}
We present \textbf{\datasetName}, a multilingual collection of datasets modeling the language a person observes from birth until they acquire a native language. We curate developmentally plausible pretraining data aiming to cover the equivalent of 100M English words of content in each of \numLanguages languages. We compile evaluation suites and train baseline models in each language. \datasetName aims to facilitate multilingual pretraining and cognitive modeling.\footnote{All code and data available through 
\href{https://babylm.github.io/babybabellm}{\tt babylm.github.io/babybabellm}.
}
\end{abstract}

\section{Introduction }

The prevailing trend in language modeling research is to prioritize scaling, both in terms of model size and training data volume \citep{kaplan2020scalinglawsneurallanguage,choshen2024hitchhikers}. 
While this approach has led to significant advances in model performance, it neglects fundamental research questions about the nature of language learning \citep{wilcox2024bigger}. 
It disincentivizes work on data-efficient modeling, which, from a practical perspective, offers benefits in terms of efficiency and accessibility. 
From a theoretical perspective, it ignores the growing mismatch between human language acquisition and language model (LM) learning. 
From infancy to maturity, English learners acquire language through exposure to less than 100M words \citep{gilkerson-etal-2017}, several orders of magnitude less than the massive pretraining corpora required by contemporary LMs surpassing 10T words \citep{bengio2025international}.

In response to the field's focus on scale, the BabyLM Challenge \citep{warstadt-etal-2023-findings} was created to redirect attention toward questions of data efficiency and developmental plausibility in language modeling. 
The shared task invites participants to propose data-efficient LMs pretrained on a fixed, developmentally plausible English corpus of child-directed speech (CDS), educational content, and other simplified texts. 
The top-performing submissions \citep{charpentier2023not,charpentier-samuel-2024-bert} have significantly improved the state of the art for models trained on the same limited data budget, even surpassing LMs trained on much larger corpora on various benchmarks.

The BabyLM Challenge has generated a new line of research on data-efficient training and cognitively-inspired modeling \citep{warstadt-etal-2023-findings, hu2024findings}, depending on the existence of developmentally plausible datasets as training corpora and supplying resources to ease and direct such research. 
However, the majority of this work has focused on English, largely due to the public availability of the pretraining corpora released for the BabyLM challenge, which is English-only. 
There is a small but growing body of work that extends the BabyLM research project beyond English \citep{salhan-etal-2024-less, shen-etal-2024-bambino, prevot-etal-2024-extending, matzopoulos-etal-2025-babylms, padovani2025childdirectedlanguagedoesconsistently,bunzeck2025construction}. 
Such efforts are crucial for developing an accurate understanding of the relationship between human language acquisition and LM learning. 
Any claim that a model is developmentally plausible can only be truly substantiated by evaluations across typologically diverse languages,  
as there is variation in acquisition trajectories between languages, and human language learning also frequently occurs in multilingual settings \citep{grosjean1989neurolinguists,slobin2014crosslinguistic,moran-2016-acqdiv,Stoll+2020+247+262}.

To facilitate this research, we create \datasetName, a multilingual collection of developmentally plausible training datasets. 
The collection includes \numLanguages languages, encompassing families primarily rooted---though not exclusively spoken---in Europe, Asia, and Africa.
For each language, we carefully select and compile publicly available datasets while prioritizing developmentally plausible data, as well as release new ones. 
This includes several categories of developmentally plausible data, such as CDS, educational resources, and other child-oriented content (e.g., books, news, and wikis aimed towards children). 
We sort languages into three tiers based on training set size, corresponding to the equivalent of respectively 100M, 10M, or 1M English words, calibrated by language-adjusted byte estimates \citep{arnett-etal-2024-bit}, to ensure comparability of data budgets across languages with differing orthographic and morphological characteristics.

To further facilitate research we also compile a list of evaluations to test models created on those domains. 
We provide a comprehensive list of existing datasets to facilitate any future questions and to provide coverage. Specifically, we cover both formal and functional competence across all languages, and include evaluation that fits the pretraining objective directly without adaptation (known as zero-shot), as well as fine-tuning based evaluation that relies on task-specific training datasets.

\setlist[itemize]{
    leftmargin=*,
    topsep=2pt,
    itemsep=1pt,
    parsep=0pt,
    partopsep=0pt,
    labelsep=4pt
}
Overall, this effort releases:
\begin{itemize}
    \item Developmentally plausible \textbf{pretraining datasets} for \numLanguages languages, collected with licenses permitting research purposes (\S\ref{section:dataset-overview}).    
    \item A \textbf{pipeline} to allow for subsequent dataset expansion with new resources and languages (\S\ref{section:data-preprocessing}). 
    \item A survey of multilingual \textbf{evaluation} tasks  (\S\ref{section:evaluation-suite}) accompanied by an evaluation suite extendable by the community. 
    \item A collection of \numLanguages monolingual \textbf{pretrained models}, 7 bilingual models and a multilingual model that we analyze in \S\ref{section:experiments}.
\end{itemize}

\section{Related Work}

The first edition of the BabyLM challenge \citep{warstadt-etal-2023-findings} released two pretraining corpora, respectively 10M and 100M words, each consisting of 39\% developmentally plausible data and a selection of high-quality corpora (e.g. Wikipedia).
The second edition \citep{hu2024findings} updated the datasets to increase the proportion of child-oriented data to 70\%.
Thus far, the BabyLM Challenge has been limited to English for training and evaluation.
In both editions, BabyLM submissions were evaluated on two types of language tasks:
1) zero-shot minimal pair challenges \citep{warstadt-etal-2020-blimp-benchmark,ivanova2024elements} benchmarking linguistic competence, world knowledge or other capabilities by testing if the model prefers a correct sentence over an incorrect one with a minor but meaningful alteration; and 2) fine-tuning based evaluations where models are further trained on a novel dataset and tested on their ability to learn the underlying task.

Beyond English, a growing body of work has begun exploring BabyLM-style models and the collection of developmentally plausible training datasets for other languages.
\citet{salhan-etal-2024-less} propose acquisition-inspired curriculum learning strategies and train small-scale LMs on age-ordered CDS for French, German, Japanese, and Chinese. 
\citet{prevot-etal-2024-extending} investigate the value of spontaneous speech corpora for BabyLM evaluation with experiments on English and French. 
\citet{matzopoulos-etal-2025-babylms} train BabyLMs for isiXhosa, a low-resource South African language, highlighting the limits of BabyLM research for languages without publicly available developmentally plausible corpora. 
\citet{capone-etal-2024-babies} release a corpus of Italian developmentally plausible training data.
\citet{padovani2025childdirectedlanguagedoesconsistently} show that training on CDS does not consistently improve grammatical learning across English, French, and German.
\citet{bunzeck2025construction} train LMs on distributionally varied subsets of a German BabyLM corpus, showing that syntax learning benefits from complex constructions while lexical learning benefits from fragmentary constructions.
Finally, \citet{shen-etal-2024-bambino} investigate developmentally plausible L2 acquisition by adapting an English BabyLM for Italian via a reward signal from a parent Italian model.
However, these works are typically forced to compile novel datasets in addition to their scientific contribution, and they do not represent a coordinated effort to compile such training data in comparable ways and across diverse languages.

Some relevant multilingual resources do exist. 
Notably, the Child Language Data Exchange System  \citep[CHILDES;][]{macwhinney2000childes} is a multilingual database of transcribed child-adult interactions, including data for over 40 languages, with varying age ranges, interaction environments, and corpus sizes. 
CHILDES serves as a starting point for most of our languages.
A previous effort to compile developmentally plausible multilingual training corpora is MAO-CHILDES \citep{yadavalli-etal-2023-slabert}, an age-ordered dataset of CHILDES corpora for five typologically diverse languages (German, French, Polish, Indonesian, and Japanese), which is used
to study cross-lingual training and L2 learning. \citet{salhan-etal-2024-less} and \citet{goriely-buttery-2025-ipa} independently release MAO-CHILDES and IPA-CHILDES for four languages (Japanese, Chinese, French, German) and a phonemized corpus based on CHILDES for 31 languages. 

\section{Dataset Creation and Overview}\label{section:dataset-overview}
The \datasetName{} dataset was created to support research on developmentally plausible language modeling across a wide range of languages. 
Our aim is to approximate the kind of linguistic input that humans are exposed to in early life, while providing clean, well-documented, high-quality data. 

\begin{figure*}[t]
    \centering
    \includegraphics[width=\textwidth]{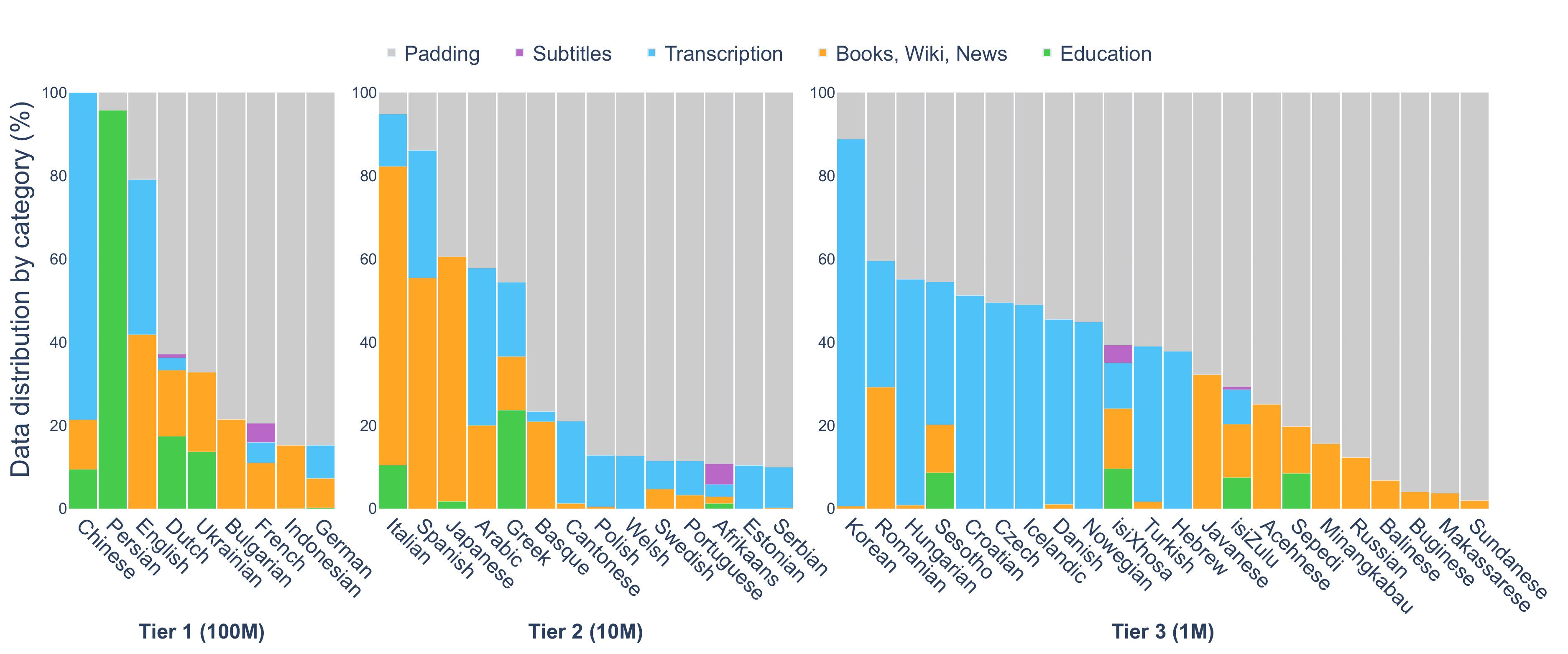}
    \caption{Training data distribution by category across languages for all data tiers in the BabyBabelLM dataset.}
    \label{fig:babylm_dataplot}
\end{figure*}

\subsection{Data Collection Principles}
The design of our datasets required various methodological choices regarding the types of data sources to include, ensuring their developmental plausibility and long-term extensibility.
In this section, we describe the criteria guiding our choices, the organizational structure of our multilingual collection, and the licensing considerations.

\subsubsection{Developmental Plausibility Criteria}
Our guiding principle in dataset construction is that of \textbf{developmental plausibility}: the idea that pretraining data should approximate the linguistic input children encounter.
To this end, we prioritized domains such as child-directed speech (CDS), educational materials, children's books, and transcribed conversations.
We deliberately excluded synthetic corpora, like TinyStories \cite{eldan2023tinystories} or TinyDialogues \cite{feng2024childdirected}, despite their developmental intention, as synthetic data has been shown to feature a reduced long tail for many linguistic measures \cite{ju-etal-2025-domain}, more uniformity in syntactic constructions \cite{Munoz-Ortiz2024-gm,strubbe2025comparison}, and less alignment with human-like discursive patterns \cite{liu-fourtassi-2025-benchmarking}. As such, it is unsuitable for our goal of approximating the full complexity of a child's linguistic environment.

In addition to content filtering, we prioritized data quality by removing noise, favoring conversational data when applicable, and standardizing the format of metadata (see Appendix~\ref{app:format}).
This preserved realism enables controlled cross-lingual comparisons, which are essential for studying the impact of linguistic variation on model learning.

\subsubsection{Community-driven Data Leadership}
\label{section:community-driven}
To ensure dataset quality, data collection for most languages was led by a researcher fluent in or familiar with that language. 
These language leads were responsible for sourcing appropriate corpora, verifying developmental plausibility, and coordinating with local experts in linguistics and 
acquisition. 

The \datasetName{} dataset is designed as a ``\mbox{living} resource''. 
As more developmentally plausible data becomes available, we aim to expand the collection both in breadth---by adding new languages---and in depth---by enriching existing ones. 
To support this, we provide an open-source pipeline that enables researchers to add entirely new languages and expand existing language datasets. 
While our initial release covers 45 languages, 16 languages rely solely on general-purpose multilingual data resources. 
We consider these entries as starting points for future, more comprehensive corpora.

We invite contributions through GitHub and Hugging Face, where researchers can submit new datasets, improvements, and evaluations. 
All additions are reviewed for compliance with our guidelines and incorporated into future versions of the dataset, ensuring proper attribution.
We hope this model of open, collaborative development will lead to broader coverage and increased utility across diverse research agendas.

\subsubsection{Licensing and Ethics}
During our data collection effort, we verified that all data is released with licenses that permit academic research, such as Creative Commons or Public Domain. When licensing information was \mbox{missing}, the right holders of each source were contacted. We release our corpus with document-level licensing information and data source attribution to ensure that each resource is used ethically and within its rights. In the rare cases where no license or contact information was available, we decided to still release the data, but under a restrictive non-commercial license.

\subsection{Dataset Composition}
Constructing a multilingual dataset that is both developmentally plausible and broadly comparable requires careful attention to how data is organized. 
Languages differ widely in the availability and type of child-relevant resources, and these differences must be accounted for without undermining cross-linguistic consistency. This section outlines how we approached this challenge, the types of data we included, and how they differ from one another. 

\subsubsection{Data categories}
\label{section:data-categories}

\paragraph{Transcription}
Children learn language mainly from spoken input, which we therefore use as our primary data source. This child-directed speech (or \textit{CDS}) differs drastically from the language data found in commonly used pretraining corpora. 
Usually, it is structurally short and simple \cite{genovese2020infantdirected}, and features high amounts of syntactic and lexical repetition \cite{tal2024infantdirected}, while its vocabulary is mostly restricted to everyday topics and children's immediate surroundings \cite{snow1977talking}. The CHILDES database 
contains a large amount of such data in the form of recorded caretaker-child interactions (e.g., during free play, meal times, or shared book reading) and manually created transcriptions. 
We used all CDS available for our target languages in CHILDES as the base of our datasets. For some languages not written in Latin script (e.g., Japanese, Greek, Persian), the CHILDES data contains transliterations, this data is excluded from our data collection. 
As children also overhear language in their surroundings, we further included as much child-available speech (adult-adult dialogue) as possible per language.


\paragraph{Education}
We included educational content aimed at children, taken from textbooks and exams, as children spend a large amount of their time in education 
and encounter this kind of input regularly. On the content level, it provides much more direct instruction than CDS, which we deem useful for our purposes. After all, our BabyBabelLMs are not only supposed to learn formal linguistic patterns from the input, but ideally also more functional (visual semantic, pragmatic, and world) knowledge.

\paragraph{Books, Wiki, News}
To approximate the whole breadth of input that children receive, we further included child-oriented media, i.e., children's books, children's wikis, child-targeted news, and other appropriate media sources. For multilingual resources, we incorporated the Ririro story collection\footnote{\url{https://ririro.com/}}, GlotStoryBooks from \citet{kargaran-etal-2023-glotlid}, and Child Wiki articles across many languages. Additionally, individual languages were enriched with monolingual resources.
In contrast to CDS, this kind of data features longer and more complex sentences \citep{cameron-faulkner2013comparison,bunzeck2025construction}, and much more diverse vocabulary and content. As such, these sources should provide a useful training signal for more complex knowledge levels, similar to educational content.

\paragraph{Subtitles}
Finally, we also used movie/TV show subtitles suitable for children. While such fictional speech does differ from natural spoken data -- for example, it features less hesitations, interjections, false starts or pauses \citep{bishop1991theres,jucker2021features,gast2023registerbased} -- it still approximates the linguistic properties of speech well and we deem it developmentally plausible. Furthermore, children are nowadays exposed to a wide variety of (video) media  \cite{gowenlock2024exposure}, and thus encounter this kind of content regularly. We also include educational content subtitles from the QED corpus \cite{abdelali-etal-2014-amara} for a small number of languages, filtering for data quality.

\paragraph{Padding}
To create comparable resources across languages, we pad our datasets to match the size of different tiers (\S\ref{sec:languages}). For padding, we use the OpenSubtitles corpora \cite{lison-tiedemann-2016-opensubtitles2016}, which are significantly larger than our other data sources. To ensure our datasets do not contain content inappropriate for children, we omitted certain categories (e.g., adult content, crime, horror). For languages where not enough OpenSubtitles data is available, we further relied on FineWeb-C and Wikipedia data, among other resources, as fallback for additional padding.

\subsubsection{Language Tiers and Coverage} \label{sec:languages}
Our dataset spans 45 languages drawn from a wide range of typological families. While Indo-European languages are well represented (22 out of 45), the collection also includes Semitic, Uralic, Bantu, Austronesian, and Sino-Tibetan languages, among others. 
This diversity was a key design goal, enabling investigation of language acquisition and modelling 
across 
distinct linguistic systems.

However, linguistic diversity is closely tied to disparities in data availability, resulting in big variations in data quantities for our set of languages.
To enable fair comparisons, we classify languages into three distinct \textbf{tiers} according to the amount of
collected data. 
Tier 1 includes languages with roughly 100 million English-equivalent tokens, Tier 2 with 10 million, and Tier 3 with 1 million. Ranking them by decreasing dataset size, the tiers contain \numTierOne, \numTierTwo, and \numTierThree languages respectively. 
This distribution  further underscores the current scarcity of developmentally plausible corpora and the need for community-driven collection efforts.


Token thresholds are \textit{calibrated} using the \textbf{byte premium} approach \cite{arnett-etal-2024-bit}, which adjusts for variation in orthographic and morphological structure by measuring the UTF-8 encoded size needed to express a fixed amount of content.
For each language, we curated as much developmentally plausible content as possible before padding to the tier threshold using fallback data sources such as OpenSubtitles. 
Figure \ref{fig:babylm_dataplot} summarizes the distribution of content categories across languages and tiers; more detailed per-language statistics are presented in  Table~\ref{tab:data_stats}.

\subsection{Data Preprocessing}
\label{section:data-preprocessing}

The data preprocessing is separated into two stages.
Initially, language-specific preprocessing was carried out by the language leads, as needed by the specific data and language (more in Appendix \ref{app:language-specific}). 
Afterwards, we apply (and release\cready{hyperref}) a uniform pipeline for all data, including standard normalization (unicode, whitespace, punctuation) and category-specific preprocessing. For dialogue transcripts, we remove linguistic annotations
For subtitle data, we remove speaker labels, music note symbols, stage directions, and timestamps. For book-like formats (educational materials, children's books, wikis) and the QED dataset, we remove XML tags and URLs. 

For language and script validation, we use GlotLID v3 \citep{kargaran-etal-2023-glotlid}. We classify sentence-like chunks of text,  created 
by splitting documents into paragraphs and 
applying sentence-based heuristics. The document's final language is assigned via a segment-based majority vote.
To maintain data quality, we filter mismatched segments within documents and discard any document that fails the overall validation. Other document metadata fields, such as the text category and license, are validated for the correct type and values when applicable (see Table~\ref{table:schema}).


\section{Evaluation Suite}
\label{section:evaluation-suite}

We create a multilingual evaluation suite that targets both the \textit{formal linguistic competence} (knowledge of linguistic rules and patterns), and the \textit{functional linguistic competence} (understanding and using language in the world) \citep{mahowald2024dissociating}.
We reviewed a large number of existing multilingual and monolingual benchmarks \cite{huang2025surveylargelanguagemodels} with the aim of ensuring all our languages have at least one evaluation dataset testing formal and one testing functional linguistic competence.

\paragraph{Formal competence}
To assess formal linguistic competence, we prioritized high-quality, language-specific minimal pair benchmarks that target a diverse set of linguistic phenomena.
This approach was applied to languages such as Basque~\citep{kryvosheieva-levy-2025-controlled}, Chinese~\citep{liu2024zhoblimpsystematicassessmentlanguage}, Japanese~\citep{someya-oseki-2023-jblimp}, 
German~\citep{vamvas-sennrich-2021-limits}, and Turkish~\citep{başar2025turblimpturkishbenchmarklinguistic}.
Where this was not possible, we employed datasets covering fewer phenomena but spanning multiple languages. In particular, for English, French, German, Russian, and Hebrew, we used CLAMS~\citep{mueller-etal-2020-cross}, a cross-lingual minimal pair benchmark built from linguist-curated templates, focusing on subject-verb number agreement. 
In our experiments we refer to the collection of these tasks as \textit{MonoBLiMP}.
Finally, we incorporated \mbox{MultiBLiMP} \citep{jumelet2025multiblimp10massivelymultilingual}, a large-scale dataset of minimal pairs automatically generated from the Universal Dependencies treebanks~\citep{nivre-etal-2017-universal}. MultiBLiMP targets subject-verb agreement in number, person, and gender, and offers the widest language coverage among our benchmarks, as detailed in Table~\ref{tab:monolingual_results}.


\paragraph{Functional competence}
We include two types of benchmarks to evaluate functional competence. The first category focuses on factual and domain-specific knowledge memorized by the model, such as Global-MMLU~\citep{singh-etal-2025-global}, INCLUDE~\citep{romanou2024includeevaluatingmultilinguallanguage}, and BMLAMA~\citep{qi-etal-2023-cross}. The second category assesses general reasoning abilities, including natural language inference, commonsense reasoning, narrative understanding, and reading comprehension. Benchmarks in this category include XNLI~\citep{conneau-etal-2018-xnli}, MultiNLI~\citep{williams-etal-2018-broad}, HellaSwag~\citep{zellers-etal-2019-hellaswag}, Belebele~\citep{bandarkar-etal-2024-belebele}, ARC~\citep{clark2018thinksolvedquestionanswering}, xstorycloze~\citep{lin-etal-2022-shot}, TruthfulQA~\citep{lin-etal-2022-truthfulqa}, XCOPA~\citep{ponti-etal-2020-xcopa}, SIB-200~\citep{adelani-etal-2024-sib}, and XWinograde~\citep{ai2:winogrande, cheng2024mubenchmultitaskmultimodalbenchmark}.
Additionally, we included XCOMPS~\citep{he2025xcompsmultilingualbenchmarkconceptual}, a multilingual conceptual minimal pair dataset with 17 languages.

\paragraph{Evaluation}
We evaluate these tasks in two ways.
Tasks that are expressed as minimal pair comparisons are evaluated using \textbf{zero-shot} prompting, based on the model's output probabilities.
For conducting these evaluations, we relied on Eleuther AI's LM Evaluation Harness \citep{eval-harness}. 
Tasks in this category are: all linguistic minimal pair tasks, XCOMPS, HellaSwag, Winogrande and XStoryCloze. For tasks involving classification and question answering we report performance after \textbf{finetuning}. The tasks on which we applied finetuning are: ARC, TruthfulQA, BMMLAMA, Belebele, INCLUDE, SIB-200, Global-MMLU, MultiNLI, XNLI and XCOPA.
We initially experimented with zero-shot prompting on these tasks as well, but the limited data size of our corpora does not allow for in-context learning mechanisms to be acquired.\footnote{\citet{olsson2022context} show that the \textit{induction heads} required for in-context learning develop only after exposure to 2.5-5 billion tokens. Developing sample-efficient methods that enable such mechanisms to emerge under much smaller data budgets, as targeted by the BabyLM challenge, is an important direction for future work but beyond the scope of this paper.}
For finetuning, we adapt the pipeline from the \textit{BabyLM Challenge 2024} evaluation framework\footnote{\url{github.com/babylm/evaluation-pipeline-2024}}. 
We limit the number of training items to a max of 8,000 items, and finetune for 10 epochs using an 80/20 train/test split.
\begin{table*}[t!]
    \setlength{\tabcolsep}{5pt}
    \renewcommand{\arraystretch}{0.9}
    \centering
    \scriptsize
\begin{tabular}{llR{0.45cm} R{0.45cm} R{0.45cm} R{0.45cm} R{0.45cm} R{0.45cm} R{0.45cm} R{0.45cm} R{0.45cm} R{0.45cm}@{\hspace{3em}}R{0.45cm} R{0.45cm} R{0.45cm} R{0.45cm} R{0.45cm} R{0.45cm}}

& \textbf{Language} & \rotatebox{45}{\textbf{sib200}} & \rotatebox{45}{\textbf{mnli}} & \rotatebox{45}{\textbf{xnli}} & \rotatebox{45}{\textbf{include}} & \rotatebox{45}{\textbf{bmlama}} & \rotatebox{45}{\textbf{global-mmlu}} & \rotatebox{45}{\textbf{arc}} & \rotatebox{45}{\textbf{truthfulqa}} & \rotatebox{45}{\textbf{belebele}} & \rotatebox{45}{\textbf{xcopa}} & \rotatebox{45}{\textbf{multiblimp}} & \rotatebox{45}{\textbf{monoblimp}} & \rotatebox{45}{\textbf{xcomps}} & \rotatebox{45}{\textbf{hellaswag}} & \rotatebox{45}{\textbf{winogrande}} & \rotatebox{45}{\textbf{xstorycloze}} \\
\midrule
 & Random & \cellcolor[rgb]{0.999,1.000,0.999} 14.3 & \cellcolor[rgb]{0.999,1.000,0.999} 33.3 & \cellcolor[rgb]{0.999,1.000,0.999} 33.3 & \cellcolor[rgb]{0.999,1.000,0.999} 25.0 & \cellcolor[rgb]{0.999,1.000,0.999} 10.0 & \cellcolor[rgb]{0.999,1.000,0.999} 25.0 & \cellcolor[rgb]{0.999,1.000,0.999} 25.0 & \cellcolor[rgb]{0.999,1.000,0.999} 25.0 & \cellcolor[rgb]{0.999,1.000,0.999} 25.0 & \cellcolor[rgb]{0.999,1.000,0.999} 50.0 & \cellcolor[rgb]{0.999,1.000,0.999} 50.0 & \cellcolor[rgb]{0.999,1.000,0.999} 50.0 & \cellcolor[rgb]{0.999,1.000,0.999} 50.0 & \cellcolor[rgb]{0.999,1.000,0.999} 25.0 & \cellcolor[rgb]{0.999,1.000,0.999} 50.0 & \cellcolor[rgb]{0.999,1.000,0.999} 50.0 \\
\end{tabular}
    \caption{
        Performance of the monolingual models trained on \datasetName. 
        All scores denote average accuracy scores, either with 0-shot prompting on the base model or on the finetuned model.
        Zero-shot performance for all tasks is provided in Table~\ref{fig:zeroshot}.
        \footnotesize
        \textsuperscript{1}For Basque we took XNLI, ARC, ThruthfulQA and XCOPA datasets from \texttt{HiTZ/xnli-eu}, \texttt{HiTZ/ARC-eu}, \texttt{HiTZ/truthfulqa-multi-MT} and \texttt{HiTZ/XCOPA-eu}.
        \textsuperscript{2}For Makasar we used a similar task to SIB200 from \texttt{nusaparagraph\_topic}. 
        \textsuperscript{3}For three South African languages (Sesotho, isiXhosa, isiZulu) we employed XNLI and Global-MMLU data from \texttt{afrixnli} and \texttt{afrimmlu}.}
        
    \label{tab:monolingual_results}
\end{table*}

\begin{figure*}
    \centering
    \includegraphics[width=0.85 \textwidth]{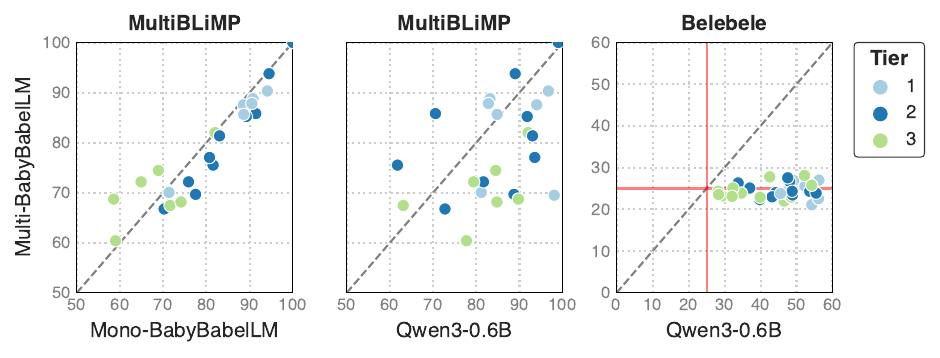}
    \caption{Language-level performance of the multilingual BabyBabelLM model against the monolingual models and Qwen3-0.6B on MultiBLiMP and Belebele.
    Each point denotes the accuracy on a specific language.
    Random performance for Belebele is denoted in red.
    }
    \label{fig:scatter}
\end{figure*}

\begin{figure}[t]
    \centering
    \includegraphics[width=\columnwidth]{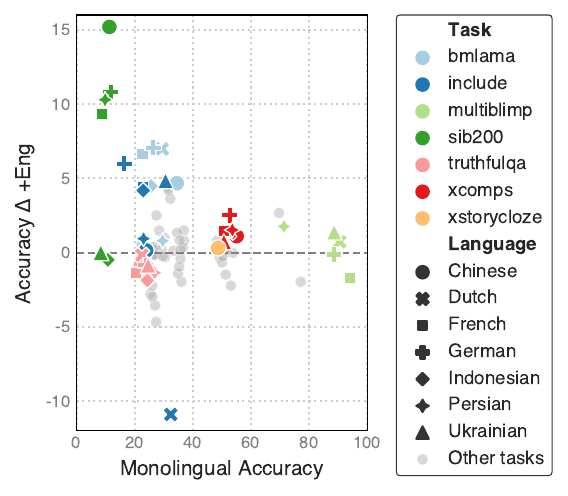}
    \caption{
    Impact of training LMs on bilingual corpora (adding English) across our evaluation suite.
    The y-axis denotes the change in accuracy from monolingual to bilingual performance.
    Dutch SIB-200 performance is omitted due to space constraints (+24.8).
    }
    \label{fig:bilingual}
\end{figure}

\section{Experiments}
\label{section:experiments}

Building on the resources outlined above, we train monolingual, bilingual and multilingual models to evaluate our benchmark suite. 

\paragraph{Setup}
For training our models, we adopt the model configurations of the GoldFish model suite \citep{chang2024goldfish}. 
For the monolingual models, we use a lightweight GPT-2 architecture with 4 transformer layers, 8 attention heads, and a hidden size of 512. 
The model uses GELU activations and standard dropout regularization (0.1) across attention, embeddings, and residual connections. 
It includes a feedforward inner dimension of 2048 and supports sequences up to 512 tokens.
For all languages, we use a BPE tokenization (trained on the training corpus), with a vocabulary size of 8,192 tokens \citep{huebner2021babyberta}.
This results in small LMs of only 17.1M parameters.
Each model is trained for 10 epochs.

For the bilingual models, we train a model using data from each language in Tier 1 and the English BabyLM (200M tokens total), keeping model configuration the same, but reducing training epochs to 5.
For the multilingual model, we increase the number of layers to 12, hidden size to 768, and vocabulary size to 32,768, accommodating the wide range of languages and scripts this model should handle.
The model is trained for only 1 epoch (around 1B tokens in total), and has 111M parameters.
We additionally compare performance against Qwen3-0.6B \citep{yang2025qwen3technicalreport}, a capable multilingual LM of modest scale.
Finally, we also experimented with training GPT-BERT \cite{charpentier-samuel-2024-bert} models on our data, the architecture that has won the BabyLM challenge of 2024.
Since these models did not outperform our GPT-2 models, we report their performance in Appendix~\ref{app:gptbert} instead.

\paragraph{Results}
The results for our monolingual models are presented in Table \ref{tab:monolingual_results}. 
Linguistic benchmarks such as MultiBLiMP yield promising results, with Tier 1 models typically scoring above 80\%.
Performance on MultiBLiMP is strongly driven by data size, with Tier 2 and 3 languages performing worse.
Performance on other benchmarks remains close to random chance (e.g., XCOPA, ARC, XCOMPS, HellaSwag). 
As such, our comparatively tiny BabyLMs only provide a starting point for further experimentation.

We further compare the results of our monolingual models for MultiBLiMP and Belebele to the multilingual BabyLM model (Multi-BabyBabelLM) and to 
Qwen3-0.6B. 
Results are summarized in Figure \ref{fig:scatter}; full results for Qwen are included in Table~\ref{tab:qwen}. 
On MultiBLiMP, the monolingual models generally outperform the multilingual one, except in four Tier 3 languages where the latter shows modest improvements. 
Compared to Qwen, results are mixed: our multilingual model is outperformed by Qwen in most cases, but remains stronger in eight languages, with no clear trend by tier.
On Belebele, both our models perform near chance, while Qwen achieves substantially higher scores in all languages. 
This pattern extends to most other benchmarks, where Qwen consistently exceeds baseline and outperforms our models on knowledge-intensive and reasoning tasks.

Figure~\ref{fig:bilingual} presents results for the bilingual models, focusing on the three best- and worst-performing tasks from the monolingual models, along with SIB-200.
All results are reported for zero-shot performance.
For several tasks---SIB-200, BMLAMA, XCOMPS, and INCLUDE---adding English as a second training language leads to consistent performance gains across most languages.
A notable exception is Dutch on INCLUDE, where bilingual training slightly reduces performance. This may be due to domain mismatch: the Dutch corpus includes high-school exam texts, and the addition of English data likely shifts the model away from this domain. 
Performance on formal linguistic tasks such as MultiBLiMP remains largely unchanged, suggesting that syntactic competence is less sensitive to bilingual input in this setup.

\section{Future Outlook}\label{sec:discuss}
\datasetName is shared research. Starting as a grassroots initiative and conducted in an open and inclusive manner, this resource was gathered by multiple experts with a shared goal. Therefore, we call for this collaboration to continue and welcome further contributions for \datasetName, even after the paper is published.
In summary, we hope that \datasetName will serve as a valuable resource for the community, facilitating reproducible, comparable, and cost-effective exploration.

To act as a complement to the resource, we provide a list of potential questions we believe this resource may aid in answering:
Do LMs acquire language more like language learners of a specific language than another?
Are there critical times for learning a second language in LMs \citep{constantinescu-etal-2025-investigating}?
Can we replicate results in studies on the border of linguistics and LLMs, where testing on only a single language might bias results \citep{arnett2025acquisition}?
Is there a way to overcome different scripts and unshared tokenizers and provide the same cross-lingual benefits between languages regardless of differences in script?
What is the right tokenization scheme across languages, and is tokenization needed at all \citep{hwang2025dynamic,rust2023language}?
While humans typically give consistent answers across languages, current LMs often do not \citep{qi-etal-2023-cross,goldman2025eclektic}. 
Even when outputs align, internal changes tend to affect only one language, indicating a degree of separation not seen in human cognition \citep{ifergan-etal-2024-identifying}. Can that be changed? 

We hope that BabyBabelLM will serve as a foundation for addressing the questions outlined above, and we invite the community to build on this resource to advance a more inclusive and systematic understanding of multilingual language acquisition and modeling.

\section*{Limitations}\label{sec:limit}
Our resources target a diverse array of audiences, and therefore our decisions are bound to not satisfy each of those perfectly. While deciding between practical constraints, data availability, and potential research needs we prioritized what we believed would make research and experimentation in the BabyLM paradigm easier. Still, we view our dataset only as a starting point. There are many more languages to be included, and even for the featured languages we imagine further untapped sources of developmentally plausible data. 

Despite our language coverage being broader than usual in NLP \citep[cf.][]{joshi-etal-2020-state}, many languages—particularly those with limited digital presence—remain underrepresented. Especially lacking are languages common in African countries and those with smaller speaker populations, which, despite our efforts, are still underrepresented in our collection. We provide instructions for submitting new languages in our GitHub
and welcome community contributions.

Although we aimed to collect as much cognitively plausible data as possible, we also want to stress that our datasets do not contain the actual language a single native speaker of any of the included languages is exposed to.
While our data approximates this input much better than standard pretraining resources (e.g., Wikipedia dumps or datasets like Dolma, \citealp{soldaini2024dolma}), the distribution of topics and formats remains only a gross approximation of the diversity experienced by a native learner.

While we calibrate dataset sizes using byte-adjusted thresholds to ensure comparability, the actual composition of developmentally plausible content varies substantially across languages. In several cases, high-quality child-directed speech (CDS) or educational material is unavailable, and we rely more heavily on fallback sources such as subtitles or Wikipedia. This variability may introduce confounds in cross-linguistic analyses and limits the strength of direct typological comparisons. We recommend that future work interpreting model differences across languages take these compositional disparities into account.

Our final limitation is the lack of cross-linguistically available  evaluation resources. Many languages are only evaluated on monolingual datasets explicitly created for them, and beyond MultiBLiMP there is currently no resource that covers all included languages. As the study of bilingualism or the acquisition of multiple languages (by models and/or humans) are intended applications of this dataset, we are also constrained by a lack of resources that explicitly target multilingual capabilities. We did not create a standardized testbed to test such questions ourselves, as they are too varied. However, we hope that our data and existing evaluations can serve as inspiration for further research in that direction.

\section*{Acknowledgments}
This work used the Dutch national e-infrastructure with the support of the SURF Cooperative using grant no. \texttt{EINF-13403}. Jaap Jumelet, Francesca Padovani and Arianna Bisazza were supported by NWO grant \texttt{VI.Vidi.221C.009}. 
Bastian Bunzeck was supported by the Deutsche Forschungsgemeinschaft (DFG, German Research Foundation) --- CRC-1646, project number 512393437, project A02.

\bibliography{anthology,custom,bastian}

\appendix
\cready{Can add this table}
\section{Author Contributions}\label{sec:contributions}
A detailed breakdown of all author contributions is provided in Table~\ref{tab:contributions}.

\begin{table*}[t]
    \centering
    \footnotesize
    \rowcolors{2}{gray!10}{white}
    \begin{tabular}{l ccccc}
    & \textbf{Paper Writing} & \textbf{Data Collection} & \textbf{Coding} & \textbf{Supervision} & \textbf{Evaluation}\\\hline
    Jaap Jumelet & $\bullet$ & \texttt{nld} & $\bullet$ & $\bullet$ & $\bullet$\\
    Abdellah Fourtassi & & \texttt{fra} & & & \\
    Akari Haga & & \texttt{jap} & & & \\
    Bastian Bunzeck & $\bullet$ & \texttt{deu} / \texttt{pol} + Wikis & & & \\
    Bhargav Shandilya & & OpenSubtitles & $\bullet$ & & \\
    Diana Galvan-Sosa & $\bullet$ & \texttt{spa} & & & \\
    Faiz Ghifari Haznitrama & $\bullet$ & Indonesian langs. + Padding & $\bullet$ & & \\
    Francesca Padovani & & \texttt{ita/spa}& $\bullet$ & & $\bullet$\\
    Francois Meyer & $\bullet$ & SA languages & & & \\
    Hai Hu & & \texttt{zho} & & & \\
    Julen Etxaniz & & \texttt{eus} & & & $\bullet$ \\
    Laurent Prévot & & \texttt{fra} & & & \\
    Linyang He & & \texttt{yue} / \texttt{zho} & & & \\
    María Grandury & & \texttt{spa} & & & \\
    Mila Marcheva & $\bullet$ & \texttt{bul} & & & \\
    Negar Foroutan & & \texttt{fas} & & & $\bullet$\\
    Nikitas Theodoropoulos & $\bullet$ & \texttt{ell} /  \texttt{ara} /  \texttt{por} & $\bullet$ & & \\
    Pouya Sadeghi & & \texttt{fas} & & & \\
    Siyuan Song & $\bullet$ & \texttt{zho} & & & \\
    Suchir Salhan & $\bullet$ & \texttt{bul} / \texttt{ron} / \texttt{spa}& $\bullet$ &   & $\bullet$ \\
    Susana Zhou & & \texttt{spa} & & & \\
    Yurii Paniv & & \texttt{ukr} & & & $\bullet$ \\
    Ziyin Zhang & & \texttt{zho} & & & \\
    Arianna Bisazza & $\bullet$ & & & $\bullet$ & $\bullet$ \\
    Alex Warstadt & $\bullet$ & & & $\bullet$ & \\
    Leshem Choshen & $\bullet$ & & & $\bullet$ & \\

    \end{tabular}
    \caption{Author contributions; language code indicates that the author was responsible for collecting and validating language-specific data for that language, as outlined in Appendix~\ref{app:language-specific}.}
    \label{tab:contributions}
\end{table*}

\section{Format considerations} \label{app:format}
In order to make our multilingual dataset easy to use and access for researchers, we format our data in a unified schema across languages consisting of self-contained documents and document-level metadata. This is applied and verified consistently for all data, including language-specific resources, multilingual datasets, and various corpora used for padding. For each document, we include details about the license and the source of the data, ensuring proper creator attribution and compliance with data sharing licenses.
Other fields encode information about a document's content, such as the text's script, target age estimate, content category, and number of tokens. The full schema of our documents and a detailed description of each field, are presented in Table~\ref{table:schema}. 


\section{Language-specific Details} \label{app:language-specific}
\subsection{Arabic}

\textbf{Dataset Description.} In recent years, there has been substantial effort towards advancing NLP for the Arabic languages. However, child-oriented resources and developmentally plausible corpora are still lacking. Some efforts have been made such as "a Compilation of an Arabic Children’s Corpus" \cite{al-sulaiti-etal-2016-compilation}, and a "leveled reading corpus of Modern Standard Arabic" \cite{al-khalil-etal-2018-leveled}. However, the data has not been made publicly available. Additionally, natural conversation and speech data, which is part of the child's linguistic environment, is often unreleased or available only under a fee.  Despite these restrictions, we present a developmentally plausible dataset for Arabic, consisting of children's books and stories, song lyrics, natural conversations, and articles from child wikis. We provide details about each data category in the paragraphs below.

Our Arabic dataset includes a large collection of different language varieties, specifically: Sudan, Egyptian, Yemeni, Meghribi, Iraqi, Levantine, Gulf and written language in Modern Standard Arabic (MSA). 
Even though spoken Arabic can vary substantially across different regions,  with speech being often mutually unintelligible, due to the scarcity of developmentally plausible data we opted to combine all the different dialects into one dataset. An important goal for the next iteration of \datasetName is to release single dialect developmentally plausible datasets for Arabic, incorporating data from recent efforts such as Atlas-Chat \cite{shang-etal-2025-atlas}. This will give us the opportunity to study the unique nature of dialects in the Arabic language, and how they interact in terms of language model training and performance in a developmentally plausible setting.

\paragraph{Books, Wiki, News.}
For books, we include the recently released Arabic Book Corpus \cite{HALLBERG2025111456}, keeping only the "children stories" category, containing both translated and original titles, mostly from the 20th century.   We also include children stories from GlotStoryBooks and Ririro and Children's Wiki articles. 

\paragraph{Transcription.}
Given that songs form a common linguistic input for children, we incorporate in our data the Habibi Corpus \cite{el-haj-2020-habibi}, consisting of song lyrics in a variety of dialects. Additionally, adult speech conversation data was collected from MagicHub\footnote{\url{https://magichub.com/datasets/}} for the Yemeni \footnote{Available here: \url{https://magichub.com/datasets/yemeni-arabic-conversational-speech-corpus/}} and the Egyptian \footnote{Available here: \url{https://magichub.com/datasets/egyptian-arabic-conversational-speech-corpus/}} dialects, as dialogue also contributes in the child's learning environment. Lastly, we include all child directed speech present in CHILDES.

\subsection{Bulgarian}

\textbf{Dataset Description.} The Bulgarian dataset is a compilation of children's literature accessed via a public library website: \url{https://chitanka.info}. The Bulgarian BabyLM corpus is the first large-scale corpus of child-appropriate Bulgarian text.

\paragraph{Books, Wiki, News.} The \texttt{Chitanka} portion of the Bulgarian BabyLM corpus consists of 28M tokens, excluding punctuation. To our knowledge, the only other similarly sourced dataset is from CHILDES, which is also included in the Bulgarian BabyLM Corpus.  The \texttt{Chitanka} library consists of several categories of books, ranging from science to literature, and has a curated section of Children and Young Adult's literature that the site owners has confirmed are free to distribute.\footnote{Excerpt from author correspondence: ``\textit{Everything in my library is completely free; you’re welcome to use any of the available resources. The books we add are supposed to be free of copyright claims. If such claims do arise—that is, if rights holders or distributors get in touch with us—those works are `quarantined' until a previously agreed period of time has elapsed.}''} \texttt{Chitanka}'s Children and Young Adult Literature sections consists of 670 texts, comprised of novels and short stories (377 texts) poems and riddles (40 texts); fairy tales (169 texts); and other children stories (25 texts); and miscellaneous children and young adult literature (68 texts).\footnote{Available here: \url{https://chitanka.info/books/category/detska-literatura}} 

\paragraph{Preprocessing.} The individual texts have been cleaned of front and back matter. Each text is provided alongside the link it was scraped from. The largest version of the dataset includes children's literature for various ages, but if one would like to restrict the dataset to a subset of earlier-age appropriate literature, this can be done by restricting the URLs which correspond to the Bulgarian Ministry of Education's programme for second and third grade summer reading (ages 6-8 years), for which the corresponding URLs are listed in the README of the dataset. A final notable detail of the dataset is that the texts are in the Cyrillic alphabet, which should be considered during preprocessing.

\textbf{Transcription.} The Bulgarian portion of CHILDES consists of 94K tokens of Child-Directed Speech (CDS) collected by \citet{Popova2020} for 5 children aged 1-2 years.

\subsection{Cantonese}

\textbf{Dataset Description.} We compile our Cantonese text corpus by consolidating four publicly available resources: the Hambaanglaang project, the GlotStory Book project, and two Cantonese datasets from CHILDES (HKU-70 Corpus and Lee/Wong/Leung Corpus).

\textbf{Books, Wiki, News.} 
\texttt{Hambaanglaang\footnote{\url{https://hambaanglaang.hk/about-us-2/}}} is an open-source repository of Cantonese graded readers created by volunteers. It offers a collection of stories designed for children across five proficiency levels, aiming to support Cantonese literacy and reading skills within the community. Detailed information about this project can be found in its official documentation.
\texttt{GlotStory Book (Hong Kong edition)\footnote{\url{https://global-asp.github.io/storybooks-hongkong}}} is a free, open-source literacy site that localizes ~40 children’s stories—originally from the African Storybook Project—into multiple languages used in Hong Kong’s “two scripts / three languages” environment (spoken \& written Cantonese, Mandarin, English). Here we only extracted Cantonese. Each story is tagged with one of five length/lexical-complexity levels and accompanied by narrated audio recordings intended to support family-, school-, and community-based language learning.
The \texttt{HKU-70 Corpus\footnote{\url{https://talkbank.org/childes/access/Chinese/Cantonese/HKU.html}}} contains 70 transcripts of interviews with 70 Cantonese-speaking children aged 2 years 6 months to 5 years 6 months. The data were collected at the University of Hong Kong and represent naturalistic child–adult interactions in preschool settings. Each child participated in a one-hour recording session, with conversations organized around familiar daily routines (e.g., bathing, dressing, feeding) to elicit a diverse range of utterances and syntactic structures. The sample was balanced by gender, and all children were prescreened using the Cantonese version of the Reynell Developmental Language Scales.
Finally, the \texttt{Lee/Wong/Leung Corpus\footnote{\url{https://talkbank.org/childes/access/Chinese/Cantonese/LeeWongLeung.html}}}, which provides longitudinal data on eight Cantonese-speaking children, each recorded for approximately one year. The recordings capture natural interactions between children, their caregivers, and occasionally other adults. Detailed metadata about the children, including their ages and family backgrounds, are included, providing valuable sociolinguistic context for the dataset.

\textbf{Preprocessing.} All datasets were cleaned to retain only complete Traditional Chinese text, with non-textual annotations such as speaker labels and syntactic tags removed. Each storybook page or conversational block is treated as a single passage-level entry. All text is tokenized using the \texttt{Qwen1.5-7B} \cite{qwen} tokenizer to ensure compatibility with downstream language modeling tasks.

\subsection{Mandarin Chinese}

\textbf{Dataset Description.} In addition to multilingual resources (CHILDES and GlotStoryBook), our Mandarin Chinese dataset includes data from multiple resources. 

\textbf{Books, Wiki, News.}
We use children's book and stories data collected from various sources. We first obtained children's stories from \texttt{Quangushi (full stories)\footnote{\url{http://quangushi.com/}}}. Then, we used children's stories from and two Chinese reading comprehension datasets: \texttt{CFT}~\citep{cui-etal-2016-consensus} and \texttt{CMRC-2019}~\citep{cui-etal-2020-cmrc2019}. These two datasets are respectively Cloze and sentence ordering benchmarks derived from children's stories. We reconstructed the complete stories using the answers provided by the authors and included the stories in our dataset. We also collected open-source children's book and children's wiki data from \texttt{WikiJunior} and \texttt{Wikibooks}.

\textbf{Education.}
For educational materials, we used several datasets that evaluate models' general knowledge through exam-style questions, as these datasets are typically well-documented and come with openly available licenses:

\begin{itemize}
    \item \texttt{GAOKAO}~\citep{zhang2023Gaokao}: an evaluation framework that uses Chinese National College Entrance Examination (GAOKAO) questions as a dataset to evaluate LLMs. The dataset includes subjective and objective questions from exams from 2010 to 2024. 
    \item \texttt{CK-12}~\citep{you2024ck12}: an evaluation for Chinese LLMs. Constructed based on multi-level knowledge graph and covers most comprehensive knowledge points in Chinese K12 field.
    \item \texttt{CSQ}~\citep{Liu2025csq}: a Chinese Science Question dataset covering four subjects and multiple topics at the Chinese primary school.
\end{itemize}
We included the full question prompts, answer choices, correct answers, and explanations. Questions in English are excluded. In addition, we collected grammatical and corrected sentences from \texttt{FCGEC}~\citep{xu2022fcgec}, a human-annotated corpus based on multi-choice
grammatical error problems.  We also collected data from a hierarchical corpus of primary school students’ compositions~\citep{zhou2024measurement}. The primary source of this corpus is elementary school student composition magazines, which ensures that the essays are of relatively high quality.
We also incorporated transcriptions in \texttt{ChildMandarin}~\citep{zhou2024childmandarin}. This dataset contains high-quality speech data collected from 397 children in China, along with carefully crafted, character-level manual transcriptions. 
We included data from \texttt{NaturalConv}~\citep{Wang_2021naturalconv}, a multi-turn topic-driven conversation dataset is also included as such conversations on daily topic are considered as children available. 

\textbf{Transcription}
Finally, we included data from Wenetspeech~\citep{zhang2022wenetspeech}, a multi-domain Mandarin corpus consisting of transcribed speech, collected from Youtube and Podcast. We only used the high-quality labeled part of the dataset, which shall be considered as children available.

\subsection{Dutch}


\textbf{Dataset Description.} The Dutch data is built from various educational sources.
Licensing laws are very strictly defined in the Netherlands, which makes it challenging to find children's literature with creative commons licenses.
Educational resources, however, are often released under CC-BY license.

\textbf{Books, Wiki, News.} 
We include the texts of all high school exams\footnote{Released publicly by \href{https://www.examenblad.nl/}{examenblad.nl}, with archives available at \href{https://www.alleexamens.nl/}{alleexamens.nl}.} from 1999 to 2024, for all Dutch high school levels: \texttt{VMBO} (age 15--16), HAVO (age 16--17), and \texttt{VWO} (age 17--18), resulting in 6.87M tokens.
We extracted 8.78M tokens from \texttt{WikiWijs\footnote{\href{https://www.wikiwijs.nl/}{wikiwijs.nl}}}, a platform for sharing educational materials by teachers for both primary and high school level.
\texttt{KlasCement\footnote{\href{https://www.klascement.net/}{klascement.net}}} provides a similar platform, focused on Flemish education, from which we extract 0.14M tokens.
Next to these educational resources, we also incorporate \texttt{BasiLex \citep{Tellings_Hulsbosch_Vermeer_vandenBosch_2014}} into the Dutch section.
\texttt{BasiLex} contains a collection of child-directed resources, extracted from children's media, children's books, and educational materials.
We collect 11.37M tokens from \texttt{BasiLex}.

\subsection{French}

\textbf{Dataset Description.} In addition to child-directed speech from CHILDES (around 4 million tokens), we include the following developmentally plausible resources, covering a range of spoken and written language that children are likely to hear or read.

\textbf{Books, Wiki, News.}
We included eighteen children's books (around 1 million tokens).\footnote{Hand‑picked from the Wikisource category: \href{https://fr.wikisource.org/wiki/Catégorie:Littérature_jeunesse}{Catégorie:Littérature jeunesse}} These were selected to match the reading level of children aged 6 to 12 and to cover a variety of story types. The collection includes classic fairy tales (Contes de Perrault, Grimm, Andersen), simple educational texts (Abécédaire du petit naturaliste, Histoires comme ça pour les petits), and famous adventure and fantasy stories like Le Tour du monde en quatre-vingts jours, L’Île au trésor, Alice au pays des merveilles, and Croc-Blanc.
Our data also includes subtitles (around 6 million tokens). This portion is made mostly of subtitles from the popular animated series Caillou, aimed at toddlers and shared on YouTube by the channel \texttt{Caillou Français – WildBrain.\footnote{\url{https://www.youtube.com/@CaillouFrench}}} It includes 1,539 video episodes. In addition to Caillou, we included other well-known children's shows in France, such as Olive et Tom (171 videos), Lou (52 videos), La vie (8 videos), and a few other youth-oriented clips (15 videos). Each subtitle document includes the YouTube video ID so the original video can be accessed. We obtained raw transcripts via the \texttt{YouTubeTranscriptApi,\footnote{\url{https://pypi.org/project/youtube-transcript-api/}}} filtered for manually entered transcripts (as opposed to automatically created ones), and fed them through the library's built-in \texttt{TextFormatter}, which strips out all timing information and reassembles each subtitle fragment as plain text.
Additionally, we included transcripts of spoken conversations (around 2 million tokens) that are not directly addressed to children, but that children could realistically overhear. We selected a number of sources from \texttt{Claire‑Dialogue‑French‑0.1\footnote{\url{https://huggingface.co/datasets/OpenLLM-France/Claire-Dialogue-French-0.1}}} \cite{openllm2023claire}, including three types of settings: spontaneous everyday conversations (in homes, cafés, or on the street), guided one-on-one interviews, and workplace meetings. The data comes from sources like \texttt{PFC\_free}, \texttt{OFROM}, \texttt{CLAPI}, \texttt{ORFEO\_coralrom}, \texttt{ParisStories}, \texttt{CFPP}, \texttt{ACSYNT}, and \texttt{ORFEO\_reunions\_de\_travail}.

\subsection{German}

\textbf{Dataset Description.} Our German data builds upon the existing German BabyLM corpus by \citet{bunzeck2025construction}, extending it with more developmentally plausible data and discarding the majority of their padding data in favor of the multilingual padding data compiled in the current project. As German is a comparatively high-resource language, we are able to supplement the multilingual resources with a variety of monolingual corpora. 

\textbf{Books, Wiki, News.} 
Five different children's wikis are available for German, including the state-sponsored \texttt{Klexikon} and \texttt{MiniKlexikon}, but also comparable efforts for Austrian German like the \texttt{Kiwithek}. In addition, we supplement this kind of educational data with the \texttt{WikiBooks Wikijunior} bookshelf, which is fairly comprehensive for German. 
As for books, we aim to make an educated selection of the \texttt{Project Gutenberg} collection featuring works for children and young adults. We include books that are considered classics of children's literature and read to this day. We further also include classics of German literature that are regularly read in middle school (e.g. works by Franz Kafka). Although they are located at the  end of the `developmentally plausible' timeline, they are plausibly encountered by many young adults in the German education system. Similarly, we also include the archives of the \texttt{Fluter} magazine, published by the German Federal Agency for Civic Education, which contains a large body of non-fiction writing aimed at adolescents and young adults. 
Moving from child-directed to child-available language, we furthermore incorporate the German section of the \texttt{CallHome corpus}~\cite{karins1997callhome}, which contains transcribed telephone conversations between adults. Such conversations could i) be plausibly overheard by children, and ii) approximate child-directed input nicely by being transcribed from spoken data, which differs quite dramatically from written data in composition (cf. \citealp{cameron-faulkner2003construction,cameron-faulkner2013comparison,bunzeck2025richness}). In a similar vein, we also incorporate the German portion of \texttt{Dreambank}~\cite{domhoff2008studying}, a large corpus of dream reports by adults and children. Despite not being originally spoken, the `self-reporting' register included in this data is closer to spoken data than ordinary writing, and social storytelling is an important component of language acquisition. Therefore, we conclude that this dataset also enhances the variety and developmental plausibility of the German data.

\subsection{Greek}

\textbf{Dataset Description.} NLP for the Greek language has developed drastically over the past few years \cite{papantoniou2024nlpgreeklanguagelonger}, with a notable example being the recent release of large language models for the Greek language: Meltemi 7B  \cite{voukoutis2024meltemiopenlargelanguage} the first such open LLM for Greek, and Krikri 8B \cite{roussis2025krikriadvancingopenlarge} further scaling up data and model sizes. Here we present, to our knowledge, the first developmentally plausible corpus for the Greek language. The data is curated as a collection of publicly available datasets,  sourced mostly from CLARIN:EL\footnote{\url{https://inventory.clarin.gr/}}, and original web-scraped children's books and stories. We present details about the dataset composition and preprocessing below. In the future we plan to include more child-directed speech data in collaboration with language acquisition researchers, incorporating efforts such as the Greek Children Spoken Language Corpus \footnote{\url{http://gcsl.ece.uth.gr/}} and the Greek-speaking Children Corpus \footnote{\url{https://gavriilidou.gr/greek-speaking-children-corpus/}}.

\paragraph{Education.} We incorporate into our data a variety of educational textbooks. We include a selection of Primary School Books \footnote{\url{https://inventory.clarin.gr/corpus/1075}} in the fields of arts, language, religion, history, and social and political sciences, aimed at grades 1-6 (ages 6-12). Apart from textbooks aimed at children, we decided to additionally include material designed for later grades and ages. Even though this content is aimed at the tail end of our target ages for developmentally plausible corpora, we consider it sufficiently relevant and representative of the linguistic input of children. Thus, we collect the CGL Modern Greek Texts corpus \footnote{\url{http://hdl.handle.net/11500/KEG-0000-0000-24FD-B}}, which comprises around 2 million words from textbooks published by the Greek Ministry of Education taught through grades 7-12 (ages 13-18) in the public school system.  We also include the corpus of Pedagogical Greek L2 textbooks \footnote{\url{ http://hdl.handle.net/11500/ATHENA-0000-0000-2631-E}}, addressed to indigenous populations or minorities learning Greek as a second language, aimed at proficiency levels A1 to C2 and ages 6-18+. Even though this resource is designed for non-native learners, we believe the material to be sufficiently close in nature to the learning resources for native Greek speakers. Finally, we include articles from Children's Wikis. 


\paragraph{Books, Wikis, News.} Numerous websites host open access children e-books and children stories for the Greek language. We identified \url{openbook.gr}\footnote{\url{https://www.openbook.gr/literature}} and \url{free-ebooks.gr}\footnote{\url{https://free-ebooks.gr/tag/16?}} as the largest such sites, and manually scraped them, selecting e-books from the categories of 
\texttt{children}, \texttt{young-adult}, and \texttt{preschool-education}. The data consists of children books in the Public Domain, as well as open access books released with permissive licenses. We also include a collection of children stories scraped from \url{paidika-paramythia.gr} \footnote{\url{https://www.paidika-paramythia.gr/16}}. The site enables any author to make a submission in collaboration with the moderators, and includes stories from tradition and  mythology, as well as original entries. Lastly, we include sort stories provided in the GlotStoryBooks corpus.  


\paragraph{Transcription.} We collect publicly available data corresponding to child-produced and child-directed speech. Child Speech \footnote{\url{http://hdl.handle.net/11500/CLARIN-EL-0000-0000-610D-5}} contains transcriptions of children's speech with a focus on narration; as the result of interviews conducted by university students with children related to them either by friendship or kinship. 
Our second addition is the Greek Student Chat Dataset \footnote{\url{http://hdl.handle.net/11500/IONION-0000-0000-5E14-1}} consists of chat among students (grades 4-18) in online  collaborative learning environments (wikis). Finally, we also include the Greek portion of CHILDES noting that speech is transcribed in the Latin script. In future efforts we plan on either removing this data or transliterating it to the Greek script.

Everyday conversations between adults are a natural stimulus for children during language development. We include in our data a corpus of written transcripts of everyday conversations between students of the Department of Linguistics \footnote{\url{http://hdl.handle.net/11500/UOA-0000-0000-5D9C-9}} that took place between 2001 and 2006. The data is further supplemented by the  Babiniotis archive \footnote{\url{http://hdl.handle.net/11500/UOA-0000-0000-2515-F}}, consisting of the same data variety recorded in 2020. The speech is authentic and idiomatic with speakers labeled, resulting in a high quality spoken Greek corpus. 


\textbf{Preprocessing.}
For the education datasets in our corpus, standard pre-processing was applied. Notably, the Primary School Books corpus required considerable cleaning and normalization efforts, containing web-scraping artifacts such as javascript code. 
 Regarding e-books, processing the text proved challenging, and required a substantial amount of manual labor. 
Initially licensing information was extracted, and the corpus was filtered 
to include only permissive licenses (e.g., cc-by-nc). For \url{openbook.gr}, license information is provided as metadata for each entry, while for \url{free-ebooks.gr} we manually annotate each book with its license as stated in the text. The stories in \url{paidika-paramythia.gr} are released as Public Domain. The text is first extracted from e-books using PyMuPDF\footnote{\url{https://github.com/pymupdf/PyMuPDF}}, and is then filtered to remove license statements, author biographies, and other information deemed irrelevant. Further document-specific normalization follows, fixing text extraction errors, removing unwanted unicode characters, and ensuring the validity of the book content. As part of this process, documents deemed unsuitable for children are excluded.
As for the transcription data, standard pre-processing was applied. Morphological and other linguistic annotations where removed from speech data. We note that to ensure anonymity, placeholders exist in conversational text that substitute real information (e.g., names, locations). 
\subsection{Italian}

\textbf{Dataset Description.} Interest in developmentally plausible NLP models for Italian has recently increased, as shown by new training setups and evaluation resources targeting child-directed-language \cite{fusco-etal-2024-recurrent,suozzi2025bambi,capone-etal-2024-babies}.
In assembling our corpus beyond the multilingual resources described in the body of the paper, we enrich it with a set of Italian-specific materials. \\

\noindent \textbf{Books, Wikis, News.} We were able to include approximately thirty books from the independent Italian publishing house Biancoenero Edizioni \footnote{\url{https://www.biancoeneroedizioni.it/}}, which kindly shared them with us upon request. The publisher has long been committed to the Alta Leggibilità (“High Readability”) project, aimed at making books accessible to all children, including those with reading difficulties. All books are written by Italian authors and are targeted at readers between the ages of 4 and 10. The themes span a range of topics including environment and ecology, bullying, mystery, diversity and inclusion, growing up and intergenerational relationships, and adventure, according to the categories listed in the publisher’s updated catalog. In addition to these recently published works, we also incorporate books from the Logos Group library \footnote{\url{https://children.logoslibrary.eu/}}. This collection comprises classic children’s stories and fairy tales authored by both Italian and foreign writers whose works are translated into Italian. The estimated target reading age for these texts ranges from approximately 6 to 14 years; however, some of these stories may be orally presented to younger children. Furthermore, we include a series of fairy tales (all from copyright expired sources) curated by the researchers in this study \cite{fusco-etal-2024-recurrent}. The book section concludes with a manually curated selection of approximately 50 titles from the Project Gutenberg catalog. These works are either explicitly included in the national curriculum for lower and upper secondary education, or authored by canonical figures whose writings are frequently excerpted in educational contexts and whose titles are broadly recognized within the Italian school system. These include both Italian and non-Italian authors. Although the language used in these works is occasionally archaic and stylistically distant from contemporary Italian,  as similarly observed in the case of German (and other languages, where applicable), their inclusion aligns with the upper boundary of the "developmentally plausible" timeline. Nevertheless, these texts remain realistically encountered by a substantial portion of young adults within the Italian educational system. 
To complement these literary sources, we also include the Italian portion of the WikiBooks Wikijunior bookshelf. This collection comprises a range of accessible entries on diverse topics (e.g., the human body, dinosaurs, the solar system), thereby extending our coverage of child-oriented reading materials beyond narrative texts. \\

\noindent \textbf{Education.} Our educational resources cover a range of materials reflecting both formal and informal learning contexts. We begin with standardized assessments, including the Italian portion of past INVALSI tests in Italian and Math at both primary and secondary levels\footnote{\url{https://www.invalsi.it/invalsi/index.php}}
. INVALSI is the national body responsible for evaluating student competencies and the quality of the education system. In addition, we incorporate an archive of national high school final examination prompts released by the Italian Ministry of Education \footnote{\url{https://www.mim.gov.it/}}
, covering the past 20 years. These standardized exams, taken by all students aged 18–19 to obtain their diploma, vary across school types but collectively represent the curricular exposure of the vast majority of Italian students and offer a representative snapshot of the competencies expected of young adults within the national system. Finally, we leverage a dataset previously curated by \citet{suozzi2025bambi} that includes children’s songs from the Zecchino D’Oro archive, a long-standing and renowned Italian music festival for children.
In addition, we include around 60 YouTube video transcripts from the animated cartoon Calimero \footnote{\url{https://www.youtube.com/@calimeroitaliano2815}}. Cartoons, while primarily designed for entertainment, also foster and support children’s language development. Our selection prioritizes episodes with consistent and realistic punctuation, filtering out automatically generated transcripts containing grammatical errors or typos.

Lastly, we complement these resources with two text simplification datasets. The first, from \citet{brunato2015design}, comprises Terence and Teacher: Terence contains 32 short children’s stories with expert-produced simplifications for readers with comprehension difficulties, while Teacher includes 18 pairs of simplified and original texts from a variety of genres (e.g., literature, textbooks). The second dataset, MultiLS, was developed for the MLSP2024 shared task \cite{shardlow2024readi} and focuses on lexical simplification. \\

\noindent \textbf{Transcription – Child Directed Speech.} We use transcripts from psycholinguistic studies on child language acquisition \cite{longobardi2015children, whittle2015insegnamento, spinelli}, which have already been employed as training data in \citet{suozzi2025bambi}. These materials originate from in vivo conversations recorded during experimental sessions and consist of utterances produced by caregivers and directed to children. \\

\noindent \textbf{Transcription – Child Available Speech.} In addition to direct caregiver input, we also consider speech that children are indirectly and routinely exposed to in their environment by overhearing adult–adult conversations. To capture this dimension, we incorporate an open-source dataset comprising 10.43 hours of transcribed conversational speech on specific topics\footnote{\url{https://magichub.com/datasets/}}
, as well as VolIP, a dataset of telephone conversations \cite{alfano2014volip} already used in \citet{fusco-etal-2024-recurrent} work.
\subsection{Japanese}

\textbf{Dataset Description.} In addition to multilingual resources, our Japanese dataset includes educational content from Wikibooks\footnote{\url{https://ja.wikibooks.org/wiki/}} and Wikijunior\footnote{\url{https://ja.wikibooks.org/wiki/Wikijunior}}, as well as children's books from Aozora Bunko\footnote{\url{https://www.aozora.gr.jp/}}.

\textbf{Books, Wiki, News.} 
For educational materials, we took data from \texttt{Wikibooks}. We used the “Elementary School Learning” section, which targets Japanese elementary school students, typically aged 6 to 12. The content covers major school subjects, including the Japanese language, social studies, mathematics, and science. We excluded pages that were still under construction, as well as those consisting primarily of numerical content(e.g., math drills). The resulting \texttt{Wikibooks} corpus contains approximately 0.2M words.
We also used \texttt{Wikijunior}, which offers educational content designed for Japanese children aged approximately 8 to 11. As with \texttt{Wikibooks}, we excluded pages that were under construction or contained only numerical content. The final \texttt{Wikijunior} corpus consists of 75 pages, totaling approximately 0.07M words from \texttt{Wikijunior}.
We complemented our data with texts from \texttt{Aozora Bunko}, a Japanese digital library that provides access to literary works in the public domain. We used the aozorabunko-clean dataset\footnote{\url{https://huggingface.co/datasets/globis-university/aozorabunko-clean}}, a cleaned version of the original collection, that includes only books whose copyrights have expired. This dataset contains storybooks, biographies, poetry, and other literary genres, with the majority being storybooks. It also contains Japanese translations of foreign literature. From this dataset, we selected only children's books. The list of children’s book titles was scraped from the category-wise list of titles on \texttt{Aozora Bunko\footnote{\url{https://yozora.main.jp/}}}. Books written in old character forms were excluded. This subset comprises 1,111 titles and totals approximately 8.7M words.
\subsection{Persian}

\textbf{Dataset Description.} Our Persian dataset includes several curated subcategories designed to support both child-centered and educational language modeling. The final collection contains about 98.5 million words across 217,880 records and consists of four parts: Children’s Books, Educational Documents, Child-Directed Speech, and Subtitles used as supplementary padding.

\textbf{Books, Wiki, News.} 
To construct a subset of educational documents, we started with \texttt{FineWeb2-HQ}~\cite{messmer2025multilingdatacomp}, a high-quality, multilingual dataset built on top of \texttt{FineWeb2}~\cite{penedo2025fineweb2pipelinescale}, as the base for our educational subset. To identify educational content within the Persian subset, we fine-tuned an XLM-R \cite{conneau-etal-2020-unsupervised} model using a regression task inspired by the \texttt{FineWeb-edu}~\cite{lozhkov2024fineweb-edu} methodology. The training data for this model were annotated using Qwen2.5-72B-Instruct~\cite{qwen2025qwen25technicalreport}, following a 5-point additive rubric designed to assess the educational suitability of a document for primary to grade school learners. The documents were then scored between 0 and 5, which were later normalized to the 0–1 range. We trained the XLM-R model to predict these normalized scores. For our final dataset selection, we applied the trained model to \texttt{Persian FineWeb2-HQ} documents. We selected documents that (1) were under 3,000 words in length (to avoid structural drift across sections), (2) had a quality score of at least 0.35 based on \texttt{FineWeb2-HQ} metadata, and (3) received a predicted educational score of 0.9 or higher. This filtering ensures that the selected documents meet at least the first four points of the rubric, which means they are coherent, suitable for grade-school learners, and contain well-structured educational material.
Our subset of children's books includes child-friendly Persian texts sourced from two main corpora: \texttt{Ririro} (Persian section) \footnote{\url{https://ririro.com/fa/}} and \texttt{GlotStoryBook}.
We scraped texts from the Persian section of the \texttt{Ririro} story collection. Although the content was mostly clean, we noticed minor inconsistencies in orthography and annotation. We applied light post-processing to fix punctuation, normalize spelling, and standardize diacritics, resulting in a clean and consistent corpus.
\texttt{GlotStoryBook} included several very short entries, such as single-word or phrase-level records. To ensure data quality and narrative coherence, we filtered out all records with fewer than three words, resulting in the removal of 123 entries from an original total of 1,150. Notably, about one-third of the \texttt{GlotStoryBook} dataset consists of “fa-diacritics” texts, which are Persian sentences written with full diacritics. These fully vocalized texts are typically used in early literacy education in Persian-speaking contexts, particularly during the first stages of primary school. They are the first form of written Persian encountered by children as they begin learning to read and write, offering a bridge toward later reading of undiacritized Persian. Their inclusion enriches the dataset with pedagogically relevant material closely aligned with actual educational practice in early schooling.

For child-directed speech, we utilized Persian transcripts from the \texttt{CHILDES} project. Notably, although the spoken language is Persian, the transcripts are written in Latin script using phonetic representations, a transcription style known as Romanized Persian. Apart from standard normalization and deduplication, no further preprocessing was applied to preserve the phonetic and linguistic characteristics of child-directed speech.

To meet our target budget of approximately 100 million words, we supplemented the dataset with Persian subtitle data. Subtitles were selected for their syntactic diversity and colloquial tone, helping to enrich the stylistic and lexical range of the dataset. The subtitles act as neutral padding rather than targeted educational or child-focused content.
\subsection{Polish}




\paragraph{Dataset Description} In addition to the multilingual resources, we add three further data sources to the Polish data. Besides these resources, we were unable to find further child-available data. Unfortunately, no spoken Polish corpora are freely available. Although some larger Polish corpora exist, projects like the National Corpus of Polish only offer rudimentary search functions and no accessible data for our purposes.

\textbf{Books, Wiki, News.} 
The \texttt{Wolne Lektury} archive contains a large number of Polish ebooks. We systematically scraped all virtual bookshelves that contain child-directed/child-available literature and included all ebooks that could be plausibly encountered by children currently learning Polish. In order to do so, we consulted native speakers of Polish and articles on classical Polish children's literature. Furthermore, we opt to include books that are translations of global children's classics (e.g. \textit{Tom Sawyer} or \textit{Alice's Adventures in Wonderland}), as they could plausibly encountered by children learning Polish. 
Besides these ebooks we also include all educational materials from the \texttt{WikiJunior} bookshelf of Polish \texttt{Wikibooks}, and educational materials from the Polish \texttt{Wikikids} website, which -- despite its name -- is not a classical wiki, but rather a general educational website.
\subsection{Portuguese}
\textbf{Dataset Description.} We present a first iteration of a developmentally plausible dataset for Portuguese. During our initial collection efforts a variety of potentially relevant resources were found, but due to time constraints have not been included in this iteration of the data.  Two such resources are PPORTAL, the Public Domain Portuguese-language Literature Dataset \cite{silva2021pportal}, and a collection of natural speech data from CORAA\footnote{\url{https://sites.google.com/view/tarsila-c4ai/coraa-versions}}.

\paragraph{Books, News, Wiki.}
Our BabyLM dataset for the Portuguese language consists primarily of sort stories from GlotStoryBooks and Ririro,  and articles from a children's wiki.

\paragraph{Transcript.}
We include child-directed speech from the Portuguese portion of CHILDES, We supplement this data with conversational spoken Brazilian Portuguese speech \footnote{\url{https://magichub.com/datasets/brazilian-portuguese-conversational-speech-corpus/}}. 
\subsection{Romanian}

\textbf{Dataset Description.} The Romanian BabyLM corpus consists of texts from CHILDES . We additionally include data from two pre-existing resources. \citet{chitez-etal-2024-towards} introduce the LEMI Romanian children’s literature corpus, which consists of 33,154 words. We also include data the children's portion of the Romanian Language Corpus collected by \citet{midrigan-ciochina-etal-2020-resources}. \footnote{\url{https://lmidriganciochina.github.io/romaniancorpus/}} The corpus consists of children's literature in its poetry and fairy tales section. 

\subsection{South African languages: Afrikaans, isiXhosa, isiZulu, Sesotho, Sepedi}

\textbf{Dataset Description.} The number of large-scale datasets and benchmarks for African languages has grown in recent years, but the African continent remains under-resourced and under-represented in NLP research \citep{ojo2025afrobenchgoodlargelanguage}. Collecting BabyLM datasets for African languages presents several challenges. Besides lacking child-directed 
speech corpora, most African languages lack even domain-general datasets of sufficient quality and scale to approximate developmentally plausible training.

As a first attempt to create BabyLM datasets for African languages, we focus specifically on the linguistically diverse context of South Africa. South Africa has 12 official languages, some of which are commonly included in massively multilingual web-scraped datasets. Importantly, all languages have some high-quality, manually curated datasets that are publicly available. This is thanks to government initiatives, such as the \texttt{South African Centre for Digital Language Resources (SADiLaR)\footnote{\url{https://repo.sadilar.org/}}}, which prioritise the development of language resources in all official languages. 
After surveying available datasets across languages, we conclude that five languages are candidates for BabyLM datasets with meaningful amount of data: Afrikaans, isiXhosa, isiZulu, Sesotho (Southern Sotho), and Sepedi (Northern Sotho). 

Afrikaans is comparably more resourced and we were able to collect a tier 2 corpus. For the other four languages, we were limited to tier 3 corpora. The proportion of data that is truly developmentally plausible varies between languages and, in some cases, falls short in comparison to higher-resourced languages. While limited in scale, our datasets demonstrate the practical feasibility of BabyLM research for low-resource languages. We hope our work serves as a starting point for future research on developmentally plausible language modelling for African languages.

\textbf{Books, Wiki, News.}
Only Afrikaans and Sesotho are represented in \texttt{CHILDES}, but we obtain child-directed and child-adult interaction corpora for all five languages from the \texttt{SA-CDI} project.\footnote{\url{https://sa-cdi.org/}} We include children's books for all five languages from the \texttt{GlotStoryBook} dataset~\citep{kargaran-etal-2023-glotlid}, originally scraped from \texttt{Nalibali\footnote{\url{https://nalibali.org/}}}, an initiative promoting children's literacy in South Africa. For educational content, we include high school exams \citep{sibeko2023highschool} for language subjects (home language and first additional language) for all five languages and \texttt{QED}~\citep{abdelali-etal-2014-amara} for Afrikaans, isiXhosa, and isiZulu. For isiXhosa, we include the descriptive sentences in \texttt{T2X}~\citep{meyer-buys-2024-triples}, a data-to-text dataset containing simplified isiXhosa sentences describing (subject, relation, object) triples in a knowledge base. 

To match the target dataset sizes (tier 2 Afrikaans and tier 3 for isiXhosa, isiZulu, Sesotho, and Sepedi), we include additional high-quality data to supplement the developmentally plausible data as needed. For Afrikaans, we include \texttt{OpenSubtitles}~\citep{lison-tiedemann-2016-opensubtitles2016}. For all five languages we include language-specific Wikipedia corpora. This still leaves us short for Sesotho and Sepedi. 
For Sepedi, we include government news articles from \texttt{Vukenzele}~\citep{lastrucci-etal-2023-preparing}. Finally, we use sentences from parallel corpora for machine translation to reach our target sizes for Sesotho and Sepedi. 
For Sepedi we include the highest quality Sepedi sentences in \texttt{WMT22}~\citep{adelani-etal-2022-findings}, as measured by language identification score. 
For Sesotho we include sentences from the \texttt{Autshumato English-Sesotho Parallel Corpus}~\citep{mckellar2022autshumato}.
\subsection{Indonesian and its local languages: Javanese, Sundanese, Balinese, Buginese, Makassarese, Minangkabau, Acehnese}

\textbf{Dataset Description.} Recent years have seen a significant increase in resources for Indonesian and its local languages, mainly due to collective efforts by \texttt{NusaCrowd}~\citep{cahyawijaya-etal-2023-nusacrowd} and \texttt{SEACrowd}~\citep{lovenia-etal-2024-seacrowd}. These initiatives have contributed a wide range of datasets, including conversational corpora, written texts, and multilingual collections. However, developmentally plausible and child-related data are still lacking. Below, we describe the data resources we found.

\textbf{Transcription.} We can only find one dataset available from the aforementioned collective efforts: \texttt{ASR-INDOCSC}, which consists of 4.5 hours of daily conversational speech from children in Indonesia, along with multilingual resources.

\textbf{Books, Wiki, News.} The main sources for cognitively and developmentally plausible data for Indonesian and its local languages come mainly from books obtained from a repository provided by the Ministry of Education \& Culture\footnote{\url{https://repositori.kemdikbud.go.id}}. These are primarily educational books and storybooks for children aged 2 to 12. Since these books are in PDF format, we used \texttt{PyPDF2}~\cite{pypdf2} and \texttt{Tesseract}~\citep{4376991} to extract their content. For data preprocessing, we use Gemma3-27B \citep{gemmateam2025gemma3technicalreport} for content filtering in three steps: filter out non-child-related books, clean and reformat the extracted book content, and then remove non-child-related content. After cleaning, \texttt{GlotLID v3}~\citep{kargaran-etal-2023-glotlid} was used for language detection and grouping, allowing data collection for Javanese, Sundanese, Balinese, Buginese, Makassarese, Minangkabau, and Acehnese. Another major source is the \texttt{Bobo children's magazine\footnote{\url{https://bobo.grid.id}}}, which contains child-targeted articles from January 2020 to May 2025, all of which are exclusively in Indonesian. In addition to these, we incorporated data from multilingual resources, specifically \texttt{GlotStoryBook}~\citep{kargaran-etal-2023-glotlid} and \texttt{Ririro\footnote{\url{https://ririro.com}}}, for Indonesian language data.

To pad the data and reach the required tiers, \texttt{OpenSubtitles}~\citep{lison-tiedemann-2016-opensubtitles2016} data were utilized for Indonesian to reach Tier 1. For local languages to reach Tier 3, we prioritized high-quality, manually curated datasets from \texttt{NusaX}~\citep{winata-etal-2023-nusax}, \texttt{NusaWrites}~\cite{cahyawijaya-etal-2023-nusawrites}, and \texttt{NusaDialogue}~\citep{purwarianti-etal-2025-nusadialogue}, followed by \texttt{Wikipedia} and \texttt{MADLAD-400}~\citep{NEURIPS2023_d49042a5} data for additional padding as needed.

\subsection{Spanish}
\textbf{Dataset Description.} As the predominant language in 21 countries, Spanish is a pluricentric language and exhibits rich diatopic variations. Far from being a homogeneous language, it encompasses a wide range of national and regional varieties, marked by distinct morphosyntactic and lexical features \cite{mayorrocher2025itssamellmsdistinguish}. As such, the term “Spanish” does not denote a single standardized form, but rather a set of linguistic norms shaped by diverse cultural and geographic contexts. The resources compiled in this dataset reflect this inherent diversity: our search for developmentally plausible materials was deliberately international, resulting in the inclusion of content from at least eight different countries.

\textbf{Books, Wiki, News.} 
A substantial portion of children’s books is sourced from the \texttt{Elejandría collection \footnote{\url{https://www.elejandria.com/coleccion/}}}, which features 19 translated bedtime stories from classical authors like Andersen, Grimm, and Perrault; 20 translated young adult classics, including "Gulliver's Travels" and "Alice in Wonderland"; and 35 original Spanish-language books by authors from Spain, Uruguay, Mexico, Nicaragua, Cuba, and Argentina, categorized under Discovering Spain and Hispanic American Literature.
Additional books were sourced from the \texttt{Logos Group} library, which granted us access upon request. This collection includes Spanish translations of well-known children’s literature, such as The Adventures of Tom Sawyer, as well as a smaller number of original Spanish texts. It also features songs, traditional Christmas carols, legends, and famous fables like The Ant and the Grasshopper.
Our dataset also includes a range of children’s stories, fairy tales, poems, traditional literature, and songs accessed via the Ministries of Education of Argentina \footnote{\url{https://www.argentina.gob.ar/educacion/historiasxleer}} and Colombia \footnote{\url{https://v1.maguared.gov.co/serie-leer-es-mi-cuento-todos-los-titulos/}}, the provincial government of Salta in Argentina \footnote{\url{https://planeamiento.edusalta.gov.ar/}}, and the educational website \textit{educ.ar.portal}~\footnote{https://www.educ.ar/}.

To capture spoken Spanish that is accessible to children, we incorporated two complementary resources.
First, we included an open-source dataset from \texttt{MagicHub\footnote{\url{https://magichub.com/datasets/}}}, comprising 5.56 hours of transcribed conversational speech in Peninsular Spanish. This dataset features 17 dialogues recorded between four pairs of speakers, covering a variety of everyday topics.
Additionally, we incorporated the \texttt{SpinTX video archive \footnote{\url{https://spintx.org/}}}, which offers curated video clips and transcripts from the Spanish in Texas Corpus. This collection of interviews with bilingual Spanish speakers residing in Texas covers a wide range of topics relevant to daily life, including family, friendship, food, culture, parenting, education, and school.
\subsection{Ukrainian}

\textbf{Dataset Description.} The Ukrainian dataset is a collection of different resources. To the best of our knowledge, there is no CHILDES-like corpus for the Ukrainian language; therefore, it has been substituted with a set of monolingual and multilingual data.

\textbf{Books, Wiki, News.} 
For the majority of developmentally plausible data, we use the \texttt{GRAC corpus}~\cite{shvedova2024plug}. This corpus consists of copyright-free texts concerning Ukraine till 1954. The dataset is heavily filtered, reducing from 100M tokens to 29M, to extract the most developmentally plausible data. First, language filtering restricts content to Ukrainian, excluding all other languages, including English, German, Russian, and others. Style-based filtering removes journalistic content, personal memoirs, religious materials, public speeches, official documents, and texts with unknown style classifications. Additionally, non-fiction works published before 1900 are excluded to maintain temporal relevance. The remaining texts are categorized into educational content (academic materials and popular science works), child-appropriate books (fiction, folklore, and poetry), and other materials (internet communication and private oral content). Additionally, we utilize the Ukrainian portion of \texttt{Wikisource}~\cite{wikisource_uk2025} as a source of fairy tales and fiction books, thereby expanding the dataset by an additional 1 million tokens.

To expand the developmentally plausible data, we incorporate the previously mentioned \texttt{GlotStorybook} and \texttt{Ririro} datasets. \texttt{Wikipedia} serves as a significant source of encyclopedic content, contributing approximately 29.1M tokens. The \texttt{FineWeb-C} corpus provides an additional 174K tokens of contemporary language use. Finally, \texttt{OpenSubtitles} contributes nearly 29.5M tokens of conversational Ukrainian text from movie and television subtitles, to which a child would most likely be exposed.

\subsection{Other Languages}
\label{appendix:other-languages}

For the rest of the languages in the \datasetName dataset, no language-specific resources were collected. Instead, these languages are populated by multilingual data resources, namely: CHILDES, GlotStoryBooks, Ririro, and Child Wikis. These languages are: \textit{Basque, Croatian, Czech, Danish, Estonian, Hebrew, Hungarian, Icelandic, Korean, Norwegian, Romanian, Russian, Serbian, Turkish, Swedish}, and \textit{Welsh} for a total of 16 out of \numLanguages languages. We welcome contributions for these, and other languages, details presented in the project website. 

\definecolor{speechcolor}{HTML}{4FC3F7}
\definecolor{educationcolor}{HTML}{47CB4D}
\definecolor{bookscolor}{HTML}{FFA724}
\definecolor{subtitlescolor}{HTML}{BA68C8}
\definecolor{paddingcolor}{HTML}{CCCCCC}

\begin{table*}[ht]
\centering

\resizebox{\textwidth}{!}{%
\begin{tabular}{llccccccccc}
\toprule
\textbf{Language} & \textbf{ISO} & \textbf{Tier} & \textbf{Byte} & \textbf{Actual Data} & \textbf{Transcription} & \textbf{Education} & \textbf{Books, Wiki,} & \textbf{Subtitles} & \textbf{Padding} & \textbf{Total} \\
& \textbf{639-3} & & \textbf{Premium} & \textbf{Size (MB)} & \textbf{Tokens} & \textbf{Tokens} & \textbf{News Tokens} & \textbf{Tokens} & \textbf{Tokens} & \textbf{Tokens} \\
\midrule
Chinese & zho & Tier 1 & 0.936 & 518.85 & 108,368,032 & 13,037,234 & 16,429,780 & 0 & 0 & 137,835,046 \\
French & fra & Tier 1 & 1.174 & 634.88 & 6,234,743 & 0 & 13,987,611 & 5,750,144 & 100,608,287 & 126,580,785 \\
Bulgarian & bul & Tier 1 & 1.812 & 981.07 & 0 & 0 & 24,799,312 & 0 & 90,563,381 & 115,362,693 \\
Indonesian & ind & Tier 1 & 1.179 & 638.87 & 17,824 & 62,188 & 17,225,662 & 0 & 96,044,750 & 113,350,424 \\
Dutch & nld & Tier 1 & 1.052 & 569.49 & 3,304,756 & 19,146,045 & 17,428,015 & 971,334 & 69,035,414 & 109,885,564 \\
German & deu & Tier 1 & 1.054 & 568.98 & 8,518,785 & 257,233 & 7,655,975 & 0 & 91,478,846 & 107,910,839 \\
\bottomrule
\end{tabular}
}
\caption{Detailed data statistics for all languages in the BabyBabelLM dataset. Tiers indicate target size equivalence to English tokens: Tier 1 (100M), Tier 2 (10M), Tier 3 (1M).}
\label{tab:data_stats}
\end{table*}

\begin{table*}[h]
\centering
\resizebox{\textwidth}{!}{%
\begin{tabular}{l@{\hskip 20pt}l@{\hskip 15pt}p{5cm}p{14cm}}

\toprule
\textbf{Field} & \textbf{Type} & \textbf{Values} &\textbf{Description}\\
\midrule
\texttt{text} & string & \textit{Una volta, c’erano}... &The content of the document.\\
\texttt{category} & string &\textbf{Transcription}\\
&&~~\textbullet~\texttt{child-directed-speech} & Speech directed to children and speech produced by children.\\
&&~~\textbullet~\texttt{child-available-speech} & Speech children are exposed to without being the target recipients (e.g., adult conversations).\\
&&\textbf{Education} &\\
&&~~\textbullet~ educational & School textbooks, exams,  and other educational material designed for children.\\
&&\textbf{Wiki, News, Books}\\
\bottomrule
\end{tabular}
}
\caption{Document-level schema for the \datasetName datasets. For each document field, we define its type, include sample values, and give a description of its use and contents. The values of the \texttt{category} field are grouped based on which high-level content category they are part of (defined in \S\ref{section:data-categories}).}
\label{table:schema}
\end{table*}

\begin{table*}[ht]
    \setlength{\tabcolsep}{5pt}
    \renewcommand{\arraystretch}{0.9}
    \centering
    \scriptsize
\begin{tabular}{llR{0.45cm} R{0.45cm} R{0.45cm} R{0.45cm} R{0.45cm} R{0.45cm} R{0.45cm} R{0.45cm} R{0.45cm} R{0.45cm} R{0.45cm} R{0.45cm} R{0.45cm} R{0.45cm} R{0.45cm} R{0.45cm} R{0.45cm}}

                                             &        \textbf{Language} &                              \rotatebox{45}{\textbf{multiblimp}} &                               \rotatebox{45}{\textbf{monoblimp}} &                                 \rotatebox{45}{\textbf{include}} &                                  \rotatebox{45}{\textbf{bmlama}} &                                \rotatebox{45}{\textbf{multinli}} &                                \rotatebox{45}{\textbf{belebele}} &                             \rotatebox{45}{\textbf{global-mmlu}} &                                     \rotatebox{45}{\textbf{arc}} &                               \rotatebox{45}{\textbf{hellaswag}} &                                    \rotatebox{45}{\textbf{xnli}} &                                   \rotatebox{45}{\textbf{xcopa}} &                             \rotatebox{45}{\textbf{xstorycloze}} &                                  \rotatebox{45}{\textbf{xcomps}} &                              \rotatebox{45}{\textbf{winogrande}} &                              \rotatebox{45}{\textbf{truthfulqa}} &                                  \rotatebox{45}{\textbf{sib200}} \\
\bottomrule
\end{tabular}
    \caption{
    Performance of Qwen3-0.6B \cite{yang2025qwen3technicalreport}. 
    All scores denote average 0-shot accuracy.
    Columns are sorted by the column order of Table~\ref{tab:monolingual_results}.
    }
    \label{tab:qwen}
\end{table*}

\begin{table*}[t!]
    \setlength{\tabcolsep}{5pt}
    \renewcommand{\arraystretch}{0.9}
    \centering
    \scriptsize
    \begin{tabular}{llR{0.45cm} R{0.45cm} R{0.45cm} R{0.45cm} R{0.45cm} R{0.45cm} R{0.45cm} R{0.45cm} R{0.45cm} R{0.45cm} R{0.45cm} R{0.45cm} R{0.45cm} R{0.45cm} R{0.45cm} R{0.45cm} R{0.45cm}}

                                             &       \textbf{Language} &                               \rotatebox{45}{\textbf{multiblimp}} &                               \rotatebox{45}{\textbf{monoblimp}} &                                  \rotatebox{45}{\textbf{bmlama}} &                                   \rotatebox{45}{\textbf{xcopa}} &                                     \rotatebox{45}{\textbf{arc}} &                                  \rotatebox{45}{\textbf{xcomps}} &                               \rotatebox{45}{\textbf{hellaswag}} &                                    \rotatebox{45}{\textbf{xnli}} &                                \rotatebox{45}{\textbf{multinli}} &                              \rotatebox{45}{\textbf{winogrande}} &                                  \rotatebox{45}{\textbf{sib200}} &                                \rotatebox{45}{\textbf{belebele}} &                             \rotatebox{45}{\textbf{global-mmlu}} &                                 \rotatebox{45}{\textbf{include}} &                             \rotatebox{45}{\textbf{xstorycloze}} &                              \rotatebox{45}{\textbf{truthfulqa}} \\
\bottomrule
\end{tabular}

    \caption{
        Zero-shot performance across all tasks of the monolingual GPT-2 models trained on \datasetName. 
        Columns are sorted by difference of the average task performance against random chance.
    }
    \label{fig:zeroshot}
\end{table*}

\section{GPT-BERT}\label{app:gptbert}
We trained monolingual GPT-BERT \citep{charpentier-samuel-2024-bert} models on all our languages in Tier 1 and 2.
Models were trained for 500 steps on Tier 1 languages (\~10 epochs) and for 250 steps on Tier 2 languages (\~25 epochs).
All models had a vocab size of 16,384, 12 layers, 768 hidden size, and 2560 intermediate size.

\begin{figure}
    \centering
    \includegraphics[width=\linewidth]{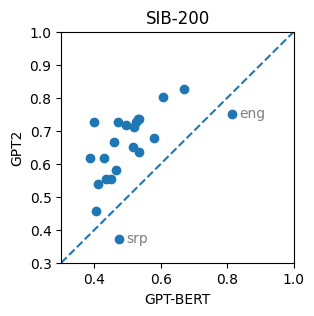}\\
    \includegraphics[width=\linewidth]{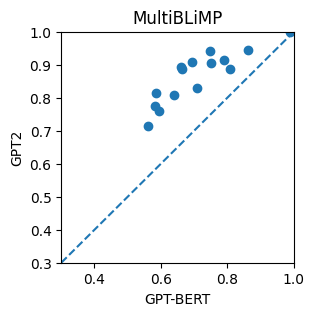}\\
    \caption{GPT-2 and GPT-BERT accuracy scores on SIB-200 and MultiBLiMP.}
    \label{fig:gptbert-results}
\end{figure}

We report the results in \ref{fig:gptbert-results}, plotting GPT-BERT against the GPT-2 models on SIB-200 and MultiBLiMP.
As can be seen, our GPT-2 models consistently outperform GPT-BERT.
We leave a more extensive exploration into finding a more optimal GPT-BERT configuration open for future work.

\end{document}